%% file: main.tex
\documentclass{article}
\usepackage[utf8]{inputenc}
\usepackage[final]{gps_mdms}
\usepackage[T1]{fontenc}    
\usepackage{hyperref}       
\usepackage{url}            
\usepackage{booktabs}       
\usepackage{amsfonts}       
\usepackage{nicefrac}       
\usepackage{microtype}      
\usepackage{xcolor}         
\usepackage{graphicx}
\usepackage{caption}
\usepackage{amsmath}
\usepackage{subcaption}
\usepackage{algorithmic}
\usepackage{algorithm}
\usepackage{amssymb}
\usepackage{xcolor}
\renewcommand{\vec}[1]{\mathbf{#1}}

\title{PACMOO-systems}
\author{%
  Alaleh Ahmadianshalchi, Syrine Belakaria, Janardhan Rao Doppa \\
  School of EECS, Washington State University \\
  \texttt{\{a.ahmadianshalchi, syrine.belakaria, jana.doppa\}@wsu.edu} \\
}

\begin{document}

\title{Preference-Aware Constrained Multi-Objective Bayesian Optimization}

\maketitle
\input{files/1-abstract}
\input{files/2-introduction}

\input{files/3-related-work}

\input{files/4-problem_setup}
\input{files/5-algorithm}
\input{files/6-experiments}
\input{files/7-conclusions}

\newpage
\footnotesize{\bibliographystyle{abbrv}}
\footnotesize{\bibliography{main}}

\end{document}

%% file: files/1-abstract.tex
\begin{abstract}

This paper addresses the problem of constrained multi-objective optimization over black-box objective functions with practitioner-specified preferences over the objectives when a large fraction of the input space is infeasible (i.e., violates constraints). This problem arises in many engineering design problems including analog circuits and electric power system design. Our overall goal is to approximate the optimal Pareto set over the small fraction of feasible input designs. The key challenges include the huge size of the design space, multiple objectives and large number of constraints, and the small fraction of feasible input designs which can be identified only after performing expensive simulations. We propose a novel and efficient preference-aware constrained multi-objective Bayesian optimization approach referred to as PAC-MOO to address these challenges. The key idea is to learn surrogate models for both output objectives and constraints, and select the candidate input for evaluation in each iteration that maximizes the information gained about the optimal constrained Pareto front while factoring in the preferences over objectives. Our experiments on two real-world analog circuit design optimization problems demonstrate the efficacy of PAC-MOO over prior methods. 

\end{abstract}

%% file: files/2-introduction.tex
\section{Introduction}

A large number of engineering design problems involve making design choices to optimize multiple objectives. Some examples include electric power systems design \cite{wang2018research,belakaria2020PSD}, design of aircrafts \cite{wang2014multi}, and design of analog circuits \cite{nicosia2008evolutionary,ACDesign}, and nanoporous materials discovery \cite{MOF}. The common challenges in such constrained multi-objective optimization (MOO) problems include the following. 1) The objective functions are unknown and we need to perform expensive experiments to evaluate each candidate design choice. 2) The objectives are conflicting in nature and all of them cannot be optimized simultaneously. 3) The constraints need to be satisfied, but we cannot evaluate them for a given input design without performing experiments. 4) Only a small fraction of the input design space is feasible. Therefore, we need to find the Pareto optimal set of solutions from the subset of feasible inputs (i.e., satisfies constraints). Additionally, in several real-world applications, the practitioners have specific preferences over the objectives. For example, the designer prefers efficiency over settling time when optimizing analog circuits.

\begin{figure*}[]
\includegraphics[width=\textwidth]{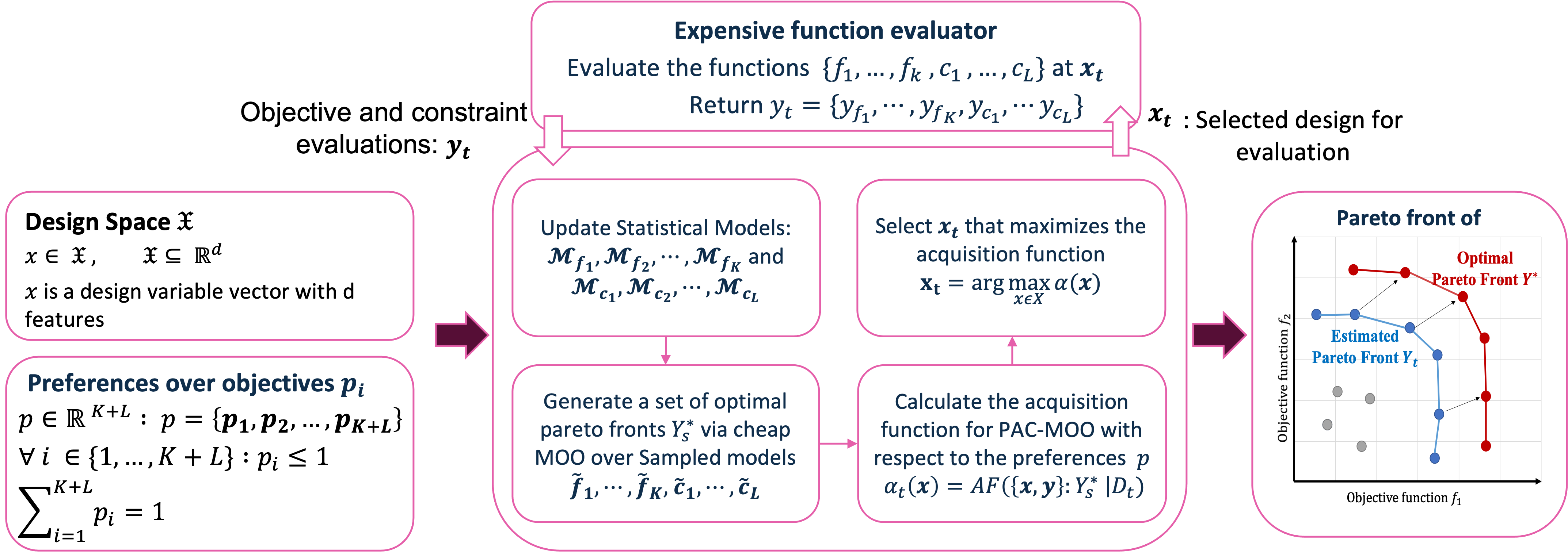}
\caption{A high-level overview of the PAC-MOO algorithm. It takes as input the input space $\mathfrak{X}$ and preferences over objectives $p$, and produces a Pareto set of candidate points as per the preferences after $T$ iterations of PAC-MOO. In each iteration $t$, PAC-MOO selects a candidate point $\vec{x}_t \in \mathfrak{X}$ to perform expensive function evaluations and the surrogate models for both objective functions and constraints are updated based on training data from the evaluated point.}\label{PACMOO_overview} 
\end{figure*}

Bayesian optimization (BO) is an efficient framework to solve black-box optimization problems with expensive objective function evaluations \cite{7352306,Jones1998}. There are no BO algorithms for simultaneously handling the challenges of black-box constraints, a large fraction of input space is invalid (doesn't satisfy all constraints), and preferences over objectives. To fill this important gap, we propose a novel and efficient information-theoretic approach referred to as {\em {\bf P}reference-{\bf A}ware {\bf C}onstrained {\bf M}ulti-{\bf O}bjective Bayesian {\bf O}ptimization (PAC-MOO)}. PAC-MOO builds surrogate models for both output objectives and constraints based on the training data from past function evaluations. PAC-MOO employs an acquisition function in each iteration to select a candidate input design for performing the expensive function evaluation. The selected input design maximizes the information gain about the constrained optimal Pareto front while factoring in the designer preferences over objectives. The experimental results on two real-world analog circuit design benchmarks demonstrate that PAC-MOO was able to find circuit configurations with higher preferred objective values (efficiency) as intended by sacrificing the overall Pareto hypervolume indicator.

\noindent {\bf Contributions.} Our key contribution is the development and evaluation of the PAC-MOO algorithm to  solve a general constrained multi-objective optimization problem. Specific contributions include: 
\begin{itemize}
    \item A tractable acquisition function based on information gain to select candidate points for performing expensive function evaluations.
    \item Approaches to increase the chances of finding feasible candidate designs and to incorporate preferences over objectives.
    \item Evaluation of PAC-MOO on two challenging analog circuit design problems and comparison with prior methods.
\end{itemize}

%% file: files/3-related-work.tex
\section{Related Work}

There are three families of approaches for solving constrained multi-objective optimization problems with expensive black-box functions. First, we can employ heuristic search algorithms such as multi-objective variants of simulated annealing \cite{vanLaarhoven1987, gielen1990analog, goodrick2021systematic}, genetic algorithms \cite{10.5555/534133,golonek2007genetic}, and particle swarm optimization \cite{Kennedy,fakhfakh2010analog,thakker2009low,vural2010component} to solve them. The main drawback of this family of methods is that they require a large number of expensive function evaluations. 
Second, Bayesian optimization (BO) methods employ surrogate statistical models to overcome the drawbacks of the previous families of approaches. The surrogate models are initialized using a small set of randomly sampled training data, i.e., input-output pairs of design parameters and objective evaluations. They are iteratively refined during the optimization process to actively collect a new training example in each iteration through an acquisition function (e.g., expected improvement). There is a large body of work on BO for single-objective optimization \cite{DU2022155,NEURIPS2019_6c990b7a}. Standard BO methods have been applied to a variety of problems including solving simple analog circuit design optimization and synthesis problems \cite{9401205,Lyu2018BatchBO,9181162,Zhang_2019,10.1145/3394885.3431543,8465872,9586172, 8116661,torun2018global}. 

Multi-objective BO (MOBO) is a relatively less-studied problem setting compared to the single-objective problem. Some of the recent work on MOBO include Predictive Entropy Search for Multi-objective Bayesian Optimization (PESMO) \cite{PESMO}, Max-value Entropy Search for Multi-Objective Bayesian optimization (MESMO) \cite{belakaria2019max}, Uncertainty-aware Search framework for Multi-Objective Bayesian Optimization (USEMO)\cite{Usemo}, Pareto-Frontier Entropy Search (PFES) \cite{suzuki2020multi}, and Expected Hypervolume Improvement \cite{daulton2020differentiable, emmerich2008computation}. Each of these methods has been shown to perform well on a variety of MOO problems. MESMO \cite{belakaria2019max} is one of the state-of-the-art algorithms that is based on the principle of output space entropy search, which is low-dimensional compared to the input space. 

Recent work extended existing approaches to the multi-objective constrained setting to account for black box constraints, notably PESMOC \cite{garrido2019predictive}, MESMOC \cite{belakaria2020max,belakaria2021output}, USEMOC \cite{belakaria2020uncertainty}. 
Existing algorithms can handle constraints that are evaluated using expensive function evaluations. However, they might not perform well when the fraction of feasible designs in the input space is small because they are hard to locate.  Additionally, none of them supports preference specifications over the output objectives. 

A parallel line of work includes several proposed approaches to incorporate preferences between different objectives \cite{abdolshah2019multi,lin2022preference}. However, simultaneously handling constraints and preferences was not well-studied. The goal of this paper is to fill this gap motivated by real-world problems in analog circuit design and electric power systems design.

%% file: files/4-problem_setup.tex
\section{Problem Setup}

\noindent {\bf Constrained multi-objective optimization w/ preferences.} Constrained MOO is the problem of optimizing $\mathbf{K} \geq 2$ real-valued objective functions $\{f_1(x), \cdots, f_{\mathbf{K}}(x)\}$, while satisfying $\mathbf{L}$ black-box constraints of the form $c_1 \geq 0, \cdots, c_\mathbf{L}(x) \geq 0$ over the given design space $\mathfrak{X}$. A function evaluation with the candidate parameters $\vec{x} \in \mathfrak{X}$ generates two vectors, one consisting of objective values and one consisting of constraint values $\vec{y} = (y_{f_1}, \cdots, y_{f_{\mathbf{K}}}, y_{c_1}, \cdots, y_{c_{\mathbf{L}}})$ where $y_{f_j} = f_j(x)$ for all $j \in \{1, \cdots, K\}$ and $y_{c_i} = C_i(x)$ for all $i \in \{1, \cdots, L\}$. We define an input vector $\vec{x}$ as feasible if and only if it satisfies all constraints. The input vector $\vec{x}$ {\em Pareto-dominates} another input vector $\vec{x'}$ if $f_j(\vec{x}) \leq f_j(\vec{x'}) \hspace{1mm} \forall{j}$ and there exists some $j \in \{1, \cdots, K\}$ such that $f_j(\vec{x}) < f_j(\vec{x'})$. 

The optimal solution of the MOO problem with constraints is a set of input vectors $\mathcal{X}^* \subset \mathfrak{X}$ such that no configuration $\vec{x'} \in \mathfrak{X} \setminus \mathcal{X}^*$ Pareto-dominates another input $\vec{x} \in \mathcal{X}^*$ and all configurations in $\mathcal{X}^*$ are feasible. The solution set $\mathcal{X}^*$ is called the optimal constrained {\em Pareto set} and the corresponding set of function values $\mathcal{Y}^*$  is called the optimal constrained {\em Pareto front}. The most commonly used measure to evaluate the quality of a given Pareto set is by calculating the Pareto hypervolume (PHV) indicator \cite{10.1145/1527125.1527138} of the corresponding Pareto front of $\vec{(y_{f_1}, y_{f_2}, \cdots, y_{f_{\mathbf{K}}}})$ with respect to a reference point $\vec{r}$. Our overall goal is to approximate the constrained Pareto set $\mathcal{X}^*$ by minimizing the total number of expensive function evaluations. When a preference specification $p$ over the objectives is provided, the MOO algorithm should prioritize producing a Pareto set of inputs that optimize the preferred objective functions.

\noindent {\bf Preferences over black-box functions.} The designer/practitioner can define input preferences over multiple black-box functions through the notion of preference specification, which is defined as a vector of scalars $\vec{p} = \{p_{f_1}, \cdots, p_{f_K}, p_{c_1}, \cdots, p_{c_L}\}$ with $0\leq p_i \leq1 $ and $\sum_{i \in \mathcal{I}} p_i = 1$ such that $\mathcal{I} = \{f_1, \cdots, f_K,c_1, \cdots, c_L\}$. Higher values of $p_i$ mean that the corresponding objective function $f_i$ is highly preferred. In such cases, the solution to the MOO problem should prioritize producing design parameters that optimize the preferred objective functions.

%% file: files/5-algorithm.tex
\section{Preference-Aware Constrained Multi-Objective Bayesian Optimization}

The general strategy behind the BO process is to employ an acquisition function to iteratively select a candidate input (i.e., design parameters) to evaluate using the information provided by the surrogate models. The surrogate models are updated based on new training examples (design parameters as input, and evaluations of objectives and constraints from function evaluations as output). 

\noindent {\bf Overview of PAC-MOO.} PAC-MOO algorithm is an instance of the BO framework, which takes as input the input space $\mathfrak{X}$, preferences over objectives $p$, expensive objective functions and constraints evaluator, and produces a Pareto set of candidate inputs as per the preferences after $T$ iterations of PAC-MOO as shown in Algorithm \ref{alg:PAC-MOO}. In each iteration $t$, PAC-MOO selects a candidate input design $\vec{x}_t \in \mathfrak{X}$ to perform a function evaluation. Consequently, the surrogate models for both objective functions and constraints are updated based on training data from the function evaluations.

\subsection {Surrogate Modeling} 

Gaussian Processes (GPs) \cite{williams2006gaussian} are suitable for solving black-box optimization problems with expensive function evaluations since they are rich and flexible models which can mimic any complex objective function. Intuitively, two candidate design parameters that are close to each other will potentially exhibit approximately similar performance in terms of output objectives. We model the objective functions and black-box constraints by independent GP models $\mathcal{GP}_{f_1}, \cdots, \mathcal{GP}_{f_K}, \mathcal{GP}_{c_1}, \cdots, \mathcal{GP}_{c_K}$ with zero mean and i.i.d. observation noise. Let $\mathcal{D} = \{(\vec{x}_i, \vec{y}_i)\}_{i=1}^{t{-1}}$ be the training data from past $t{-1}$ function evaluations, where $\vec{x}_i \in \mathfrak{X}$ is a candidate design and $\vec{y}_i = \{y_{f_1}^i, \cdots, y_{f_K}^i, y_{c_1}^i, \cdots, y_{c_L}^i\}$ is the output vector resulting from evaluating the objective functions and constraints at $\vec{x}_i$.

\subsection{Acquisition Function}
\label{appendix_acq}

The state-of-the-art MESMO approach for solving MOO problems \cite{belakaria2019max} proposed to select the input that maximizes the information gain about the optimal {\bf Pareto front} for evaluation. However, this approach did not address the challenge of handling black-box constraints which can be evaluated only through expensive function evaluators. To overcome this challenge, MESMOC \cite{belakaria2020max} maximizes the information gain between the next candidate input for evaluation $\vec{x}$ and the optimal constrained Pareto front $\mathcal{Y}^*$:

\begin{align}
        \alpha(\vec{x}) &= I(\{\vec{x}, \vec{y}\}, \mathcal{Y}^* \mid D) = H(\vec{y} \mid D, \vec{x}) - \mathbb{E}_{\mathcal{Y}^*} [H(\vec{y} \mid D,   \vec{x}, \mathcal{Y}^*)] \label{eqn_symmetric_mesmo1}
\end{align}

In this case, the output vector $\vec{y}$ is $K+L$ dimensional: $\vec{y}$ = $(y_{f_1}, y_{f_2},\cdots,y_{f_K},y_{c_1} \cdots y_{c_L})$ where $y_{f_j} = f_j(x) \forall j \in \{1,2, \cdots, K\}$ and $y_{c_i}$ = $C_i(x)\forall i \in \{1,2, \cdots, L\}$.
Consequently, the first term in Equation (\ref{eqn_symmetric_mesmo1}), entropy of a factorizable $(K+L)$-dimensional Gaussian distribution $P(\vec{y}\mid D, \vec{x})$, can be computed in closed form as shown below:

\begin{align}
    H(\vec{y} \mid D, \vec{x}) = \frac{(K+C)(1+\ln(2\pi))}{2} +  \sum_{j = 1}^K  \ln (\sigma_{f_j}(\vec{x}))+  \sum_{i = 1}^L  \ln (\sigma_{c_i}(\vec{x})) \label{eqn_unconditioned_entropy1}
\end{align}

where $\sigma_{f_j}^2(\vec{x})$ and  $\sigma_{c_i}^2(\vec{x})$ are the predictive variances of $j^{th}$ function and $i^{th}$ constraint GPs respectively at input $\vec{x}$. The second term in Equation (\ref{eqn_symmetric_mesmo1}) is an expectation over the Pareto front $\mathcal{Y}^*$. We can approximately compute this term via Monte-Carlo sampling as shown below: 

\begin{align}
 \mathbb{E}_{\mathcal{Y}^*} [H(\vec{y} \mid D, \vec{x}, \mathcal{Y}^*)] \simeq \frac{1}{S} \sum_{s = 1}^S [H(\vec{y} \mid D, \vec{x}, \mathcal{Y}^*_s)] \label{eqn_summation}
\end{align}

where $S$ is the number of samples and $\mathcal{Y}^*_s$ denote a sample Pareto front. There are two key algorithmic steps to compute this part of the equation: 1) How to compute constrained Pareto front samples $\mathcal{Y}^*_s$?; and 2) How to compute the entropy with respect to a given constrained Pareto front sample $\mathcal{Y}^*_s$? We provide solutions for these two questions below.

\hspace{2.0ex} {\bf 1) Computing constrained Pareto front samples via cheap multi-objective optimization.} To compute a constrained Pareto front sample $\mathcal{Y}^*_s$, we first sample functions and constraints from the posterior GP models via random Fourier features \cite{PES,random_fourier_features} and then solve a cheap constrained multi-objective optimization over the $K$ sampled functions and $L$ sampled constraints.

\hspace{2.5ex}{\em Cheap MO solver.} We sample $\Tilde{f}_i$ from GP model $\mathcal{GP}_{f_j}$ for each of the $K$ functions and $\Tilde{C}_j$ from GP model $\mathcal{GP}_{c_j}$ for each of the $L$ constraints. A {\em cheap} constrained multi-objective optimization problem over the $K$ sampled functions  $\Tilde{f}_1,\Tilde{f}_2,\cdots,\Tilde{f}_k$ and the $L$ sampled constraints  $\Tilde{C}_1,\Tilde{C}_2,\cdots,\Tilde{C}_L$ is solved to compute the sample Pareto front $\mathcal{Y}^*_s$. Note that we refer to this optimization problem as cheap because it is performed over sampled functions and constraints, which are cheaper to evaluate than performing expensive function evaluations. We employ the popular constrained NSGA-II algorithm \cite{deb2002nsga,deb2002fast} to solve the constrained MO problem with cheap sampled objective functions and constraints.

\hspace{2.0ex}{\bf 2) Entropy computation with a sample constrained Pareto front.}
Let $\mathcal{Y}^*_s = \{\vec{v}^1, \cdots, \vec{v}^l \}$ be the sample constrained Pareto front,  where $l$ is the size of the Pareto front and each $\vec{v}^i$ is a $(K+L)$-vector evaluated at the $K$ sampled functions and $L$ sampled constraints $\vec{v}^i = \{v^i_{f_1},\cdots,v^i_{f_K},v^i_{c_1},\cdots,v^i_{c_L}\}$. The following inequality holds for each component $y_j$ of the $(K+L)$-vector $\vec{y} = \{y_{f_1},\cdots,y_{f_K},y_{c_1},\cdots y_{c_L}\}$ in the entropy term $H(\vec{y} \mid D,   \vec{x}, \mathcal{Y}^*_s)$:

\begin{align}
 y_j &\leq \max \{v^1_j, \cdots v^l_j \} \quad \forall j \in \{f_1,\cdots,f_K,c_1,\cdots,c_L\} \label{inequality1}
\end{align}

The inequality essentially says that the $j^{th}$ component of $\vec{y}$ (i.e., $y_j$) is upper-bounded by a value obtained by taking the maximum of $j^{th}$ components of all $l$ $(K+L)$-vectors in the Pareto front $\mathcal{Y}^*_s$. This inequality had been proven by a contradiction for MESMO \cite{belakaria2019max} for all objective functions $j \in \{f_1,\cdots,f_K\}$. We assume the same  for all constraints $j \in \{c_1,\cdots,c_L\}$.

By combining the inequality (\ref{inequality1}) and the fact that each function is modeled as an independent GP, we can approximate each component $y_j$ as a truncated Gaussian distribution since the distribution of $y_j$ needs to satisfy $ y_j \leq \max \{v^1_j, \cdots v^l_j \}$. Let $y_s^{c_i*} = \max \{v^1_{c_i}, \cdots v^l_{c_i} \}$ and $y_s^{f_j*} = \max \{v^1_{f_j}, \cdots v^l_{f_j} \}$. Furthermore, a common property of entropy measure allows us to decompose the entropy of a set of independent variables into a sum over entropies of individual variables \cite{information_theory}:

\begin{align}
H(\vec{y} \mid D,   \vec{x}, \mathcal{Y}^*_s) = \sum_{j=1}^K H(y_{f_j}|D, \vec{x}, y_s^{f_j*}) +\sum_{i=1}^L H(y_{c_i}|D, \vec{x}, y_s^{c_i*})  \label{eqn_sep_ineq1}
\end{align}
The r.h.s is a summation over entropies of  $(K+L)$-variables $\vec{y} = \{y_{f_1},\cdots,y_{f_K},y_{c_1},\cdots y_{c_L}\}$.
The differential entropy for each $y_j$ is the entropy of a truncated Gaussian distribution \cite{entropy_handbook} and is given by the following equations:

\begin{align}
H(y_{f_j} \mid D, \vec{x}, y_s^{f_j*}) \simeq [\frac{(1 + \ln(2\pi))}{2}+  \ln(\sigma_{f_j}(\vec{x})) +  \ln \Phi(\gamma_s^{f_j}(\vec{x})) - \frac{\gamma_s^{f_j}(\vec{x}) \phi(\gamma_s^{f_j}(\vec{x}))}{2\Phi(\gamma_s^{f_j}(\vec{x}))}] \label{eqn_entropy_fj}
  \end{align}

\begin{align}
H(y_{c_i}|D, \vec{x}, y_s^{c_i*}) \simeq 
[\frac{(1 + \ln(2\pi))}{2}+  \ln(\sigma_{c_i}(\vec{x})) +  \ln \Phi(\gamma_s^{c_i}(\vec{x})) - \frac{\gamma_s^{c_i}(\vec{x}) \phi(\gamma_s^{c_i}(\vec{x}))}{2\Phi(\gamma_s^{c_i}(\vec{x}))}]    
    \label{eqn_entropy_cj}
\end{align} 

Consequently, we have: 
\begin{align}
    & H(\vec{y} \mid D,   \vec{x}, \mathcal{Y}^*_s) \simeq \nonumber
    \sum_{j=1}^K \left[\frac{(1 + \ln(2\pi))}{2}+  \ln(\sigma_{f_j}(\vec{x})) +  \ln \Phi(\gamma_s^{f_j}(\vec{x})) - \frac{\gamma_s^{f_j}(\vec{x}) \phi(\gamma_s^{f_j}(\vec{x}))}{2\Phi(\gamma_s^{f_j}(\vec{x}))}\right] \nonumber\\
& \hspace{2.0cm} + \sum_{i=1}^L \left[\frac{(1 + \ln(2\pi))}{2}+  \ln(\sigma_{c_i}(\vec{x})) +  \ln \Phi(\gamma_s^{c_i}(\vec{x})) - \frac{\gamma_s^{c_i}(\vec{x}) \phi(\gamma_s^{c_i}(\vec{x}))}{2\Phi(\gamma_s^{c_i}(\vec{x}))}\right]    
    \label{eqn_entropy_closed1}
\end{align}

where $\gamma_s^{c_i}(x) = \frac{y_s^{c_i*} - \mu_{c_i}(\vec{x})}{\sigma_{c_i}(\vec{x})}$, $\gamma_s^{f_j}(x) = \frac{y_s^{f_j*} - \mu_{f_j}(\vec{x})}{\sigma_{f_j}(\vec{x})}$, 
$\phi$ and $\Phi$ are the p.d.f and c.d.f of a standard normal distribution respectively. By combining equations (\ref{eqn_unconditioned_entropy1}) and (\ref{eqn_entropy_closed1}) with equation (\ref{eqn_symmetric_mesmo1}), we get the final form of our acquisition function as shown below:
\begin{align}
\alpha(\vec{x}) \simeq \sum_{i\in \mathcal{I}} AF(i,x) ~\text{with} ~ i \in \mathcal{I} ~\text{and} ~ \mathcal{I}=\{c_1 \cdots c_L,f_1 \cdots f_K\} \label{eqn_final1} 
\end{align} 

And 
\begin{equation}
    AF(i,x)= \sum_{s=1}^S \frac{\gamma_s^{i}(\vec{x}) \phi(\gamma_s^{i}(\vec{x}))}{2\Phi(\gamma_s^{i}(\vec{x}))} - \ln \Phi(\gamma_s^{i}(\vec{x})) \label{component_mes}
\end{equation}

\subsection{Convex Combination for Preferences} \label{convex_preference}

We now describe how to incorporate preference specification (when available) into the acquisition function. The derivation of the acquisition function proposed in Equation \ref{eqn_final1} resulted in a function in the form of a summation of an entropy term defined for each of the objective functions and constraints as $AF(i,x)$. Given this expression, the algorithm will select an input while giving the same importance to each of the functions and constraints. However, as an example, in problems such as circuit design optimization, efficiency is typically the most important objective function. The designer would like to find a trade-off between the objectives. Nevertheless, candidate circuits with high voltage and very low efficiency might be useless in practice. Therefore, we propose to inject preferences from the designer into our algorithm by associating different weights to each of the objectives. A principled approach would be to assign appropriate preference weights resulting in a convex combination of the individual components of the summation $AF(i,x)$. Let $p_i$ be the preference weight associated with each individual component. The preference-based acquisition function is defined as follows (see Algorithm 2): 
\begin{align} \label{AF_final}
    &\alpha_{pref}(\vec{x}) \simeq \sum_{i\in \mathcal{I}} p_i\times AF(i,x) ~\text{with} ~ i \in \mathcal{I} & s.t \sum_{i\in \mathcal{I}} p_i =1
\end{align}
It is important to note that in practice if a candidate design does not satisfy the constraints, the optimization will fail regardless of the preferences over objectives. Therefore, the cumulative weights assigned to the constraints have to be at least equal to the total weight assigned to the objective functions:
\begin{align}
 \sum_{i\in \{c_1\cdots c_L\}} p_i =  \sum_{i\in \{f_1, \cdots, f_K\}} p_i =\frac{1}{2}
\end{align}
Given that satisfying all the constraints is equally important, the weights over the constraints would be equal. Finally, only the weights over the functions will need to be explicitly specified. 

\begin{algorithm}[h]
\caption{PAC-MOO Algorithm}
\label{alg:PAC-MOO}
\scriptsize
\small{Inputs}: {Input space $\mathcal{X}$, black-box functions $\{f_1,...,f_{\mathbf{K}}\}$, constraint functions $\{c_1,...,c_L\}$, preferences $\vec{p}=\{p_{f_1},\cdots,p_{f_K},p_{c_1},\cdots,p_{c_L}\}$, number of initial points $\mathcal{N}_0$, number of iterations $T$}\
\begin{algorithmic}[1] 
\STATE Initialize Gaussian processes for functions $\mathcal{M}_{f_1},\cdots,\mathcal{M}_{f_\mathcal{K}}$ and constraints  $\mathcal{M}_{c_1},\cdots, \mathcal{M}_{c_\mathcal{L}}$ by evaluating them on $\mathcal{N}_0$ initial design parameters
\FOR{ each iteration t = $\mathcal{N}_0$ to T+$\mathcal{N}_0$ }

\IF{feasible design parameters $\vec{x}_{feasible} \notin \mathcal{D}$} 
\STATE Select design parameters $\vec{x}_{t} \leftarrow \arg max_{\vec{x}\in \mathcal{X}} \hspace{2 mm} \alpha_{prob}(\vec{x}) $ \textit{\# eq. \ref{alpha_prod}}
\ELSE
\STATE Select design parameters $\vec{x}_{t} \leftarrow \arg max_{\vec{x}\in \mathcal{X}} \hspace{2 mm} \alpha_{pref}(\vec{x},\vec{p}) $ in Algorithm \ref{alg:pefMESMOC} \\
    \qquad \qquad \qquad \qquad \qquad \qquad \textbf{s.t} $(\mu_{c_1}\geq 0, \cdots,\mu_{c_L}\geq 0 )$
\ENDIF
\STATE Perform function evaluations using the selected design parameters \\  
\qquad $\vec{x}_{t}$: $\vec{y}_{t} \leftarrow (f_1(\vec{x}_{t}),\cdots,f_K(\vec{x}_{t}),C_1(\vec{x}_{t}),\cdots,C_L(\vec{x}_{t}))$ 
\STATE Aggregate data: $\mathcal{D} \leftarrow \mathcal{D} \cup \{(\vec{x}_{t}, \vec{y}_{t})\}$ 
\STATE Update models $\mathcal{M}_{f_1},\cdots, \mathcal{M}_{f_K}$ and $\mathcal{M}_{c_1},\cdots, \mathcal{M}_{c_L}$ using $\mathcal{D}$
\ENDFOR
\STATE \textbf{return} the Pareto set of feasible design parameters from $\mathcal{D}$
\end{algorithmic}
\end{algorithm}

\begin{algorithm}[H]
\caption{Preference based Acquisition function ($\alpha_{pref}$)}\label{alg:pefMESMOC}
\small
\begin{flushleft}
\textbf{$\alpha_{pref}(\vec{x},\vec{p})$ }
\end{flushleft}
\begin{algorithmic}[1] 

\FOR{Each sample $s \in \{1,\cdots,S\}$} 
\STATE Sample functions $\Tilde{f}_j \sim \mathcal{M}_{f_j}, \quad \forall{j \in \{1,\cdots, K\}} $
\STATE  Sample constraints $\Tilde{C}_i \sim \mathcal{M}_{c_i}, \quad \forall{i \in \{1,\cdots, L\}} $
\STATE  Solve {\em cheap} MOO over $(\Tilde{f}_1, \cdots, \Tilde{f}_K)$ {constrained by} $(\Tilde{C}_1, \cdots, \Tilde{C}_L)$\\
\qquad  \qquad  $\mathcal{Y}_s^* \leftarrow \arg max_{x \in \mathcal{X}} (\Tilde{f}_1, \cdots, \Tilde{f}_K)$ \\ \qquad \qquad \textbf{s.t} $(\Tilde{C}_1\geq 0, \cdots, \Tilde{C}_L\geq 0)$
\ENDFOR
\FOR{$i \in \mathcal{I} ~\text{and} ~ \mathcal{I}=\{c_1 \cdots c_L,f_1 \cdots f_K\}$} 
\STATE Compute $AF(i,x)$ based on $S$ samples of $\mathcal{Y}_s^*$ via Equation \ref{component_mes}
\ENDFOR
\STATE \textbf{Return $\sum_{i\in \mathcal{I}} p_i\times AF(i,x)$} 
\end{algorithmic}
\end{algorithm}

\noindent {\bf Finding Feasible Regions of Design Space}

The acquisition function defined in equation \ref{AF_final} will build constrained Pareto front samples $\mathcal{Y}_s^*$ by sampling functions and constraints from the Gaussian process posterior. The posterior of the GP is built based on the current training data $\mathcal{D}$. The truncated Gaussian approximation defined in Equations \ref{eqn_entropy_fj} and \ref{eqn_entropy_cj} requires the upper bound $y_s^{f_j*}$ and $y_s^{c_i*}$ to be defined. 
However, in the early Bayesian optimization iterations of the algorithm, the configurations evaluated may not include any feasible design parameters. This is especially true for scenarios where the fraction of feasible design configurations in the entire design space is very small. In such cases, the sampling process of the constrained Pareto fronts $\mathcal{Y}_s^*$ is susceptible to failure because the surrogate models did not gather any knowledge about feasible regions of the design space \textit{yet}. Consequently, the upper bounds $y_s^{f_j*}$ and $y_s^{c_i*}$ are not well-defined and the acquisition function in  \ref{AF_final} is not well-defined.
Intuitively, the algorithm should first aim at identifying feasible design configurations by maximizing the probability of satisfying all the constraints. We define a special case of our acquisition function for such challenging scenarios as shown below: 

\begin{align}
    \alpha_{prob}(x)=\prod_{i=1}^{L} Pr(c_i(x)\geq0) \label{alpha_prod}
\end{align}

This acquisition function enables an efficient feasibility search due to its exploitation characteristics \cite{gardner2014bayesian}. Given that the probability of constraint satisfaction is binary (0 or 1), the algorithm will be able to quickly prune unfeasible regions of the design space and move to other promising regions until it identifies feasible design configurations. This approach will enable a more efficient search over feasible regions later and accurate computation of the acquisition function. The complete pseudo-code of PAC-MOO is given in Algorithm \ref{alg:PAC-MOO}. 

%% file: files/6-experiments.tex
\section{Experimental Setup and Results}

In this section, we present experimental evaluation of PAC-MOO and baseline methods on two challenging analog circuit design problems.

\noindent\textbf{Baselines.} We compare PAC-MOO with state-of-the-art constrained MOO evolutionary algorithms, namely, NSGA-II \cite{deb2002fast} and MOEAD \cite{4358754}. We also compare to the constrained MOO method, the Uncertainty aware search framework for multi-objective Bayesian optimization with constraints (USEMOC) \cite{belakaria2020uncertainty}. We evaluated two variants of USEMOC: USEMOC-EI and USEMOC-TS, using expected improvement (EI) and Thompson sampling (TS) acquisition functions.

\noindent\textbf{PAC-MOO}: We employ a Gaussian process (GP) with squared exponential kernel for all our surrogate models. We evaluated several preference values for the efficiency objective function.
PAC-MOO-0 refers to the preference being equal over all objectives and constraints. PAC-MOO-1 refers to assigning 80\% preference to the efficiency objective and equal importance to other functions and constraints, resulting in a preference value $p_i=0.5 \times 0.8=0.4$ for the efficiency. With PAC-MOO-2, we assign a total preference of 85\% to the objective functions with 92\% importance to the efficiency resulting in a preference value of $p_i=0.85 \times 0.92=0.782$. We assign equal preference to all other functions.  With PAC-MOO-3, we assign more importance to the objective functions by assigning a total of 0.65 preference to them and 0.35 to the constraints. Additionally, we provide 88\% importance to the efficiency resulting in a preference value of $p_i=0.65 \times 0.88=0.572$.

\noindent\textbf{Evaluation Metrics}: The \textit{Pareto Hypervolume (PHV)} indicator is a commonly used metric to measure the quality of the Pareto front \cite{phd/dnb/Zitzler99}. PHV is defined as the volume between a reference point and the Pareto front.  After each circuit simulation, we measure the PHV for all algorithms and compare them. To demonstrate the efficacy of the preference-based PAC-MOO, we compare different algorithms using the maximum efficiency of the optimized circuit configurations as a function of the number of circuit simulations.

\noindent {\bf Benchmarks: } { \em 1.Switched-Capacitor Voltage Regulator (SCVR) design optimization setup.}  
The constrained MOO problem for SCVR circuit design consists of 33 input design variables, nine objective functions, and 14 constraints. 
Every method is initialized with 24 randomly sampled circuit configurations. 
{\em 2. High Conversion Ratio (HCR) design setup.} 
The constrained MOO problem for HCR circuit design consists of 32 design variables, 5 objective functions, and 6 constraints.
Considering that the fraction of feasible circuit configurations in the design space is extremely low (around 4\%), every method is initialized with 32 initial feasible designs provided by a domain expert.

In all our preference-based experiments, we assign a preference value to the efficiency objective and assign all other black-box functions (the rest of the objectives and the constraints) equal preference.

It is noteworthy that neither evolutionary algorithms nor the baseline BO method USEMOC are capable of handling preferences over objectives. This is an important advantage of our PAC-MOO algorithm, which we demonstrate through our experiments.

\begin{figure}[ht!]
\begin{subfigure}{.5\textwidth}
\includegraphics[width=\textwidth]{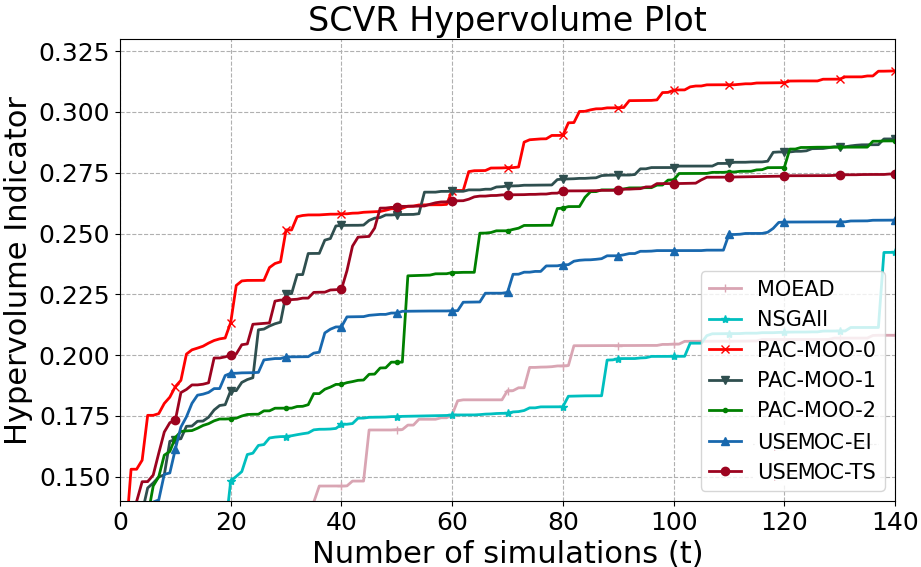}
\caption{Hypervolume - SCVR}\label{scvr_hv_plot}
\end{subfigure}
\begin{subfigure}{.5\textwidth}
\includegraphics[width=\textwidth]{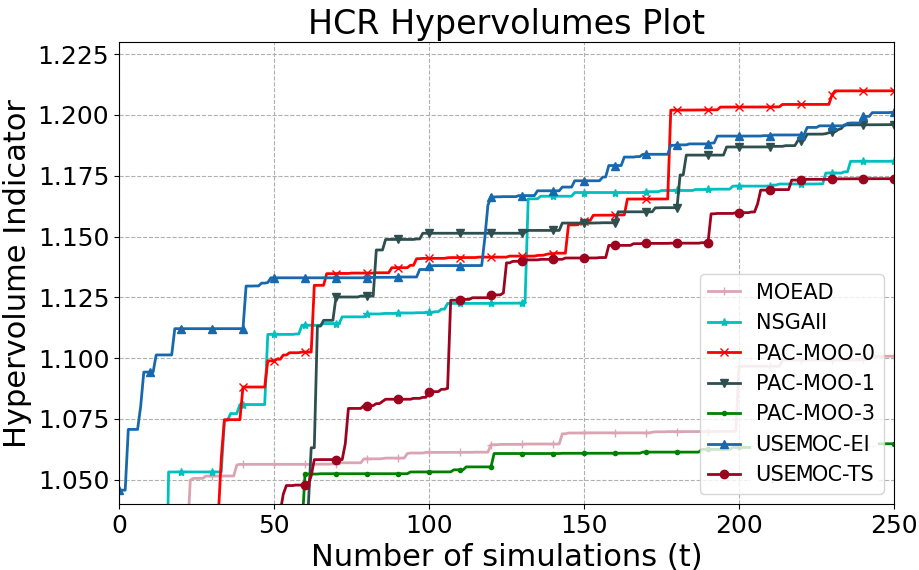}
\caption{Hypervolume - HCR}\label{hcr_hv_plot}
\end{subfigure}
\begin{subfigure}{.5\textwidth}
\includegraphics[width=\textwidth]{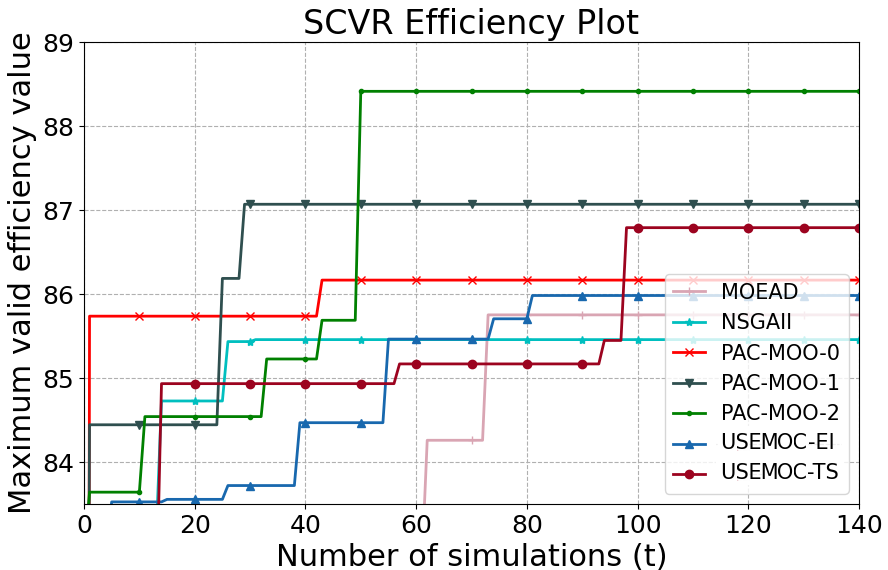}
\caption{Efficiency - SCVR}\label{scvr_max_efficiency_plot}
\end{subfigure}
\begin{subfigure}{.5\textwidth}
\includegraphics[width=\textwidth]{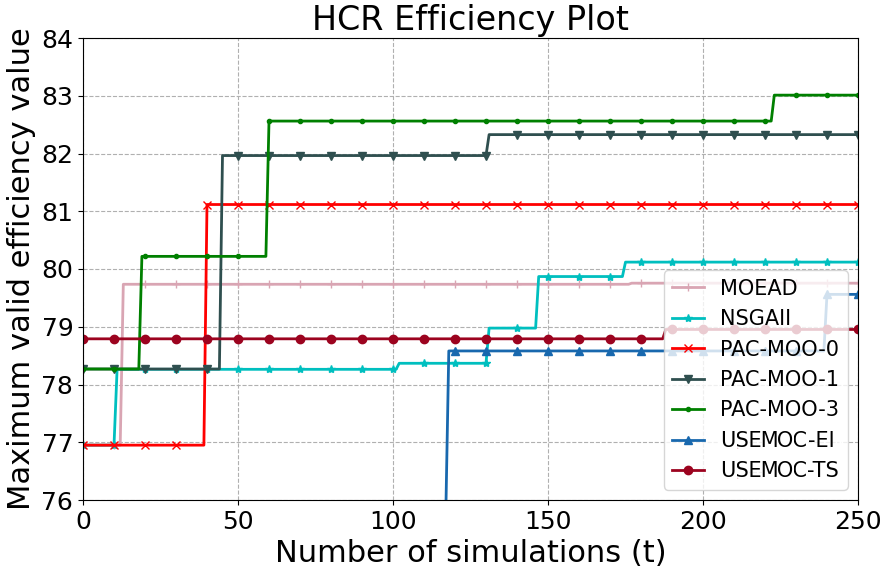}
\caption{Efficiency - HCR}\label{hcr_max_efficiency_plot}
\end{subfigure}
\caption{Hypervolume and Efficiency of optimized circuits with preferences vs. No of simulations}
\end{figure}

\noindent {\bf Hypervolume of Pareto set vs. No of circuit simulations.} Figures \ref{scvr_hv_plot} and \ref{hcr_hv_plot} show the results for PHV of Pareto set as a function of the number of circuit simulations for SCVR and HCR design, respectively. An algorithm is considered relatively better if it achieves higher hypervolume with a lower number of circuit simulations. 
We make the following observations. {\bf 1)} PAC-MOO with no preferences (i.e., PAC-MOO-0) outperforms all the baseline methods. This is attributed to the efficient information-theoretic acquisition function and the exploitation approach to finding feasible regions in the circuit design space. {\bf 2)} At least one version of USEMOC performs better than all evolutionary baselines: USEMOC-EI for both SCVR and HCR designs. These results demonstrate that BO methods have the potential for accelerating analog circuit design optimization over evolutionary algorithms .
{\bf 3)} The performance of PAC-MOO with preference (i.e., PAC-MOO-1,2,3) is lower in terms of the hypervolume since the metric evaluates the quality of general Pareto front, while our algorithm puts emphasis on specific regions of the Pareto front via preference specification. This behavior is expected, nevertheless, we notice that the PHV with  PAC-MOO-1 and PAC-MOO-2 is still competitive and degrades only when a significantly high preference is given to efficiency (PAC-MOO-3). 

\noindent {\bf Efficiency of optimized circuits with preferences.} Since efficiency is the most important objective for both SCVR and HCR circuits, we evaluate PAC-MOO by giving higher preference to efficiency over other objectives. Figures \ref{scvr_max_efficiency_plot} and \ref{hcr_max_efficiency_plot} show the results for maximum efficiency of the optimized circuit configurations as a function of the number of circuit simulations for SCVR and HCR design optimization. {\bf 1)} As intended by design, PAC-MOO with preferences outperforms all baseline methods, including PAC-MOO without preferences. {\bf 2)} The improvement in maximum efficiency of uncovered circuit configurations for PAC-MOO with preferences comes at the expense of loss in hypervolume metric as shown in Figure \ref{scvr_hv_plot} and Figure \ref{hcr_hv_plot}.  

%% file: files/7-conclusions.tex
\section{Summary}

Motivated by challenges in hard engineering design optimization problems (e.g., large design spaces, expensive simulations, a small fraction of configurations are feasible, and the existence of preferences over objectives), this paper proposed a principled and efficient Bayesian optimization algorithm referred to as PAC-MOO. The algorithm builds Gaussian process based surrogate models for both objective functions and constraints and employs them to intelligently select the sequence of input designs for performing experiments. The key innovations behind PAC-MOO include a scalable and efficient acquisition function based on the principle of information gain about the optimal constrained Pareto front; an effective exploitation approach to find feasible regions of the design space; and incorporating preferences over multiple objectives using a convex combination of the corresponding acquisition functions. Experimental results on two challenging analog circuit design optimization problems demonstrated that PAC-MOO outperforms baseline methods in finding a Pareto set of feasible circuit configurations with high hyper-volume using a small number of circuit simulations. With preference specification, PAC-MOO was able to find circuit configurations that optimize the preferred objective functions better. 

\vspace{1.0ex}

\noindent {\bf Acknowledgements.} The authors gratefully acknowledge the support from National Science Foundation (NSF) grants IIS-1845922, OAC-1910213, and SII-2030159. The views expressed are those of the authors and do not reflect the official policy or position of the NSF.